\documentclass{article}

\PassOptionsToPackage{numbers, compress}{natbib}

\usepackage[accepted]{icml2025}
\usepackage{amsmath}
\usepackage{graphicx}
\usepackage{float}
\usepackage{amssymb}
\usepackage{algorithm}
\usepackage{algorithmic}
\usepackage{subcaption}
\usepackage{makecell}
\usepackage{multirow}
\usepackage{booktabs}
\usepackage{enumitem}

\usepackage{natbib}
\usepackage[utf8]{inputenc} 
\usepackage[T1]{fontenc}    
\usepackage{hyperref}       
\usepackage{url}            
\usepackage{booktabs}       
\usepackage{amsfonts}       
\usepackage{nicefrac}       
\usepackage{microtype}      
\usepackage{xcolor}         
\usepackage{setspace}

\makeatletter
\renewcommand{\normalsize}{\fontsize{9}{11}\selectfont}
\renewcommand{\footnotesize}{\fontsize{7}{9}\selectfont}
\makeatother

\title{Do LLMs Understand Limit Order Book Dynamics?}

\author{%
  Junxiao Chen, Paul Glasserman \\
  Columbia Business School \\
  \texttt{\{J1Chen27, pg20\}@gsb.columbia.edu} \\
}
\date{}

\begin{document}

\maketitle

\begin{abstract}

A large language model (LLM) trained on synthetic limit order book (LOB) data achieves near perfect scores in generating valid sequences of LOB events. However, the LLM's implicit world model fails to learn the state of the LOB. This deficiency leads to biased estimates and spurious predictability in using the LLM to forecast future LOB events. Our analysis uses novel tests of an LLM's world model, extending prior work from deterministic settings to the stochastic dynamics needed for the LOB.
    
\end{abstract}

\section{Introduction}

If a large language model (LLM) learns to recognize and generate valid sequences of limit order book (LOB) events --- order arrivals, execution, and cancellations --- has it developed a correct world model of the LOB? Does such an understanding matter?
These are the questions we investigate.

We train a transformer-based LLM from scratch using events in a simulated LOB. The LLM performs extremely well in generating valid sequences and in finding sequences that move the LOB from a starting state to a target state. By these measures, the LLM appears to understand key concepts underlying an LOB: transactions can only occur at the best bid or ask price; market orders can only execute against opposite side resting orders; a cancellation order requires the presence of a limit order to be canceled.

At the same time, we find evidence of systematic errors in the LLM's implicit world model of the LOB. A correct understanding of the LOB's dynamics becomes important in applying the trained LLM to a task related to but different from the task on which it was trained.

In our setting, the related task is forecasting. We have in mind a trader or market-maker using the trained LLM to predict, e.g., whether the next order will be a buy or a sell order. We find consistent evidence of spurious predictability: the LLM finds predictability where none exists, and thus delivers misleading forecasts. This is where the abstract concept of an incorrect world model takes on practical consequences.

By using simulated data, we are able to ensure that the state of the data-generating LOB is a Markov chain. 
Given the current state, the past history of the LOB is therefore irrelevant to its future evolution, yet the LLM's forecasts are influenced by past events. Our framework thus evaluates the LLM under the null hypothesis of no predictability. Events in an actual LOB may well exhibit some predictability, but an LLM's ability to provide reliable forecasts is suspect if it finds predictability under the null. More fundamentally, its implicit world model of an LOB appears to be flawed.

Prior work on inferring and evaluating LLMs' world models includes \cite{hazineh2023linear,
li2023emergent,toshniwal2022chess} in the setting of games, \cite{guan2023leveraging} in task planning, \cite{liutransformers,vafa2024evaluating} in 
deterministic finite automata (DFA),
and \cite{vafa2025has} in the setting of physical laws. In these cases, the ``true'' model is deterministic whereas the evolution of an LOB is inherently stochastic.
We adapt and apply some tests studied in the general framework of \cite{vafa2024evaluating}.

Our focus on biased forecasts and spurious predictability is new and particularly relevant to the financial setting. We introduce new tests for this property. Our {\it kernel-level total variation (TV)} measures deviations between an LLM's next-token probabilities and the true distribution kernel. Our {\it history-level total variation (TV)} measures the erroneous influence of past events on an LLM's forecast distribution. We also apply a regression test to evaluate spurious predictability or miscalibration in the probabilities of future events using information on past events.

From a practical perspective, our results indicate that valid-sequence tests are insufficient in training an LLM for downstream tasks (such as forecasting events) in LOB. Larger training datasets or new training methods are needed to improve an LLM's world model of an LOB. Our results use small LOB settings to make training from scratch feasible; the issues we document would likely be more severe with a larger state space.
A growing literature focuses on generating large synthetic financial datasets; see, e.g.,
\cite{assefa2020generating,kong2024large,NagyFSLCZF23}.

Section~\ref{s:lob} reviews LOBs and formulates our LOB dynamics. Section~\ref{s:data_genaration} constructs the datasets.
Section~\ref{s:training} describes our model training process. 
Section~\ref{s:empirical_baseline} introduces a simple baseline for comparison.
Section~\ref{s:diagnostics} uses existing diagnostics to evaluate our models.
Section~\ref{s:forecasts} introduces and applies our new tests. 
Section~\ref{s:scale_up} tests larger LOB settings.
Section~\ref{s:conclusion} concludes.

\section{Limit Order Book}
\label{s:lob}

\subsection{States}

\begin{figure}[H]
  \centering
 \includegraphics[trim=177.57pt 279.24pt 264.45pt 148.62pt, clip, width=0.7 \linewidth]{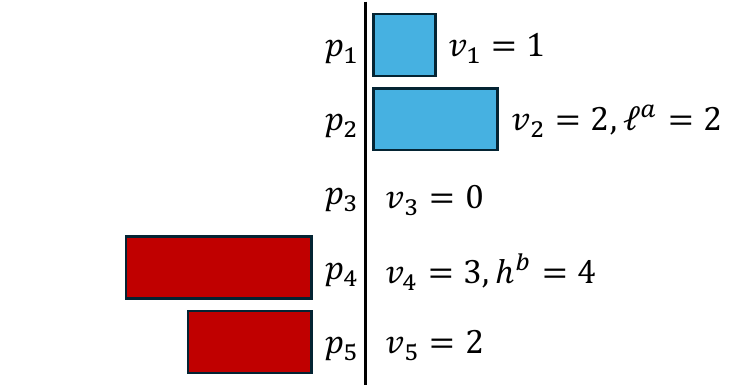}
  \caption{Representation of a limit order book}
  \label{f:lob}
\end{figure}

Figure~\ref{f:lob} illustrates a limit order book. This example has five fixed price levels $p_1>\cdots>p_5$. There are $v_1=1$ orders to sell at price $p_1$ and $v_2=2$ orders to sell at $p_2$; there are $v_4=3$ orders to buy at price $p_4$ and $v_5=2$ orders to buy at $p_5$. 
The notation $\ell^a=2$ indicates that the lowest asking price is $p_2$, and similarly $h^b=4$ indicates that the highest bid price is $p_4$. 

More generally, we represent the state of the LOB by the vector $(\ell^a,h^b,v_1,\dots,v_P)$,
where $P$ is the fixed number of possible price levels. 
For $i=1,\dots,\ell^a$, $v_i$ is the quantity of sell orders asking $p_i$. For $i=h^b,\dots,P$, $v_i$ is the quantity of buy orders bidding $p_i$.
If there are no bids (buy orders), then
$h^b=P+1$, and if there are no ask (sell) orders then $\ell^a=0$.

Any price levels ($p_3$ in the figure) between the bids at the bottom and the asks at the top constitute the spread region and should have $v_i=0$. To keep the state space finite, we cap the total number of orders at any price level at $V_1$, and the size of each individual order at $V_2$. (See, e.g., \cite{cartea2015algorithmic} for general background on limit order books.)

\subsection{Events and State Transitions}
\label{s:events}

The state of the LOB evolves through the arrival of limit orders, cancellation orders, and market orders. We associate a token with each possible event. The token  \texttt{ask\_p\_i\_v} represents the arrival of an ask order to sell $v$ shares at price $p_i$. A limit buy order is similarly written as \texttt{bid\_p\_i\_v}. The cancellation orders \texttt{w\_ask\_p\_i\_v} and \texttt{w\_bid\_p\_i\_v} remove $v$ units from the corresponding side at price $p_i$. A market sell order \texttt{m\_ask\_v} consumes $v$ units from the bid side starting from the best bid $p_{h^b}$; a market buy order \texttt{m\_bid\_v} consumes $v$ units from the ask side starting from the best ask $p_{\ell^a}$.

Limit orders follow standard matching logic. If an arriving limit buy order bids at a price $p_i$ below the lowest ask price $p_{\ell^a}$, the order rests in the book at price $p_i$, provided that the resulting volume does not exceed $V_1$. If instead the submitted limit buy order bids at $p_i \geq p_{\ell^a}$ (a crossing bid), it executes against the opposite side starting from the lowest asking price $p_{\ell^a}$. If the submitted size is larger than the available sell volume at prices less than or equal to the bid $p_i$, the unfilled residual rests at $p_i$. The arrival of limit sell orders works analogously. Cancellation orders are valid only when the corresponding side and price level currently contain at least the requested cancellation volume. Market orders are valid only if the opposite side has enough total volume to fully execute the order. 

The rules of the LOB determine the set of valid events $\mathcal{A}(s)$ in each state $s$. Each valid event $a\in \mathcal{A}(s)$ deterministically maps the LOB state $s$ to a new state $s' = T(s,a)$, as in a deterministic finite automaton (DFA), the framework used in \cite{vafa2024evaluating}.
There may be more than one event that moves the state from $s$ to $s'$. 

Once we specify the probability $\mu(a|s)$ of each event $a$ in each state $s$, the evolution of the state becomes a Markov chain. The transition probability from $s$ to $s'$ is given by
$$
K(s,s') = \sum_{a\in\mathcal{A}(s):T(s,a)=s'}\mu(a|s).
$$
The resulting finite-state Markov chain is irreducible and aperiodic, and therefore admits a unique stationary distribution \(\pi\), which satisfies \(\pi K = \pi\). For simplicity, we assume that, in each LOB state, all valid events (tokens) are equally likely, $\mu(a|s) = 1/|\mathcal{A}(s)|$. We have obtained consistent results using non-uniform kernels that assign higher probabilities to buy or sell orders, inducing upward or downward price pressure.

\section{Synthetic Datasets Construction}
\label{s:data_genaration}

We associate a state ID token with every LOB state, and, as described above, we associate a token with every event. 
Each training sequence is represented as
\begin{equation}
s_{\mathrm{start}} \;\; s_{\mathrm{goal}} \;\; a_1 \;\; a_2 \;\; \cdots \;\; a_T \;\; \texttt{end},
\label{seq}
\end{equation}
where $s_{\mathrm{start}}$ is the initial state ID, $s_{\mathrm{goal}}$ is a target state ID, and $a_1,\ldots,a_T$ are valid LOB events. The special token \texttt{end} is valid only when current state equals the target state. 
We train an LLM to respond to a prompt of the form
$(s_{\mathrm{start}}, s_{\mathrm{goal}})$ by generating a sequence of valid event tokens that move the LOB from $s_{\mathrm{start}}$ to $s_{\mathrm{goal}}$.

This format requires the LLM to learn which sequences of events are valid from any LOB state $s_{\mathrm{start}}$. But it goes further in requiring the model to learn how to reach a target state $s_{\mathrm{goal}}$, which suggests an ability to plan or anticipate the impact of multiple events and thus a deeper understanding of the LOB's transition structure.%
\footnote{The goal state acts as a conditioning signal, analogous to contextual conditioning in standard language modeling, requiring the model to learn multi-step reachability rather than only local transition probabilities. This does not conflict with our subsequent prediction tests, since this conditioning is averaged out over goal states drawn from the same data-generation process.}

Following \cite{vafa2024evaluating},
we generate two types of datasets of sequences \eqref{seq}: random walks and shortest paths.

For the random walk dataset, we draw $s_{\mathrm{start}}$ from the stationary distribution $\pi$ of the LOB Markov chain. We generate $a_1$ from $\mu(\cdot|s_{\mathrm{start}})$, the next-token distribution in the current state.
The event $a_1$ moves the LOB to a new state $s'$.
With probability $p_{\text{stop}}$, 
the sequence terminates and we set $s_{\mathrm{goal}}=s'$.
With probability $1-p_{\text{stop}}$, we generate $a_2$ from $s'$, and the process repeats. This construction gives the sequence length $T$ a geometric distribution with mean $1/p_{\text{stop}}$.
The memoryless property of the geometric distribution implies that the number of events observed thus far along a sequence carries no information about how many more events will be observed before the sequence terminates.
In our experiments, we cap the maximum sequence length to keep the length bounded.

In the shortest-path dataset, we require that \eqref{seq} move from the start to the goal state with the fewest possible transitions $T$. 
Details of the construction are given in Appendix~\ref{s:shortestpath}.

The random walk dataset is intended to approximate observational data. Given a history of events in an actual LOB, one could randomly divide the historical sequence into segments, record the start and end states for each segment, and train an LLM to generate similar segments. The shortest-path dataset is for comparison; we expect the shortest paths to reveal more about the LOB dynamics but to explore the state space less thoroughly.

\section{LLM Training}
\label{s:training}

For each dataset, we train a decoder-only transformer model on token sequences of the form \eqref{seq}. We use a GPT-style causal decoder architecture with 12 transformer layers, embedding dimension 768, and 12 attention heads, giving approximately 86.4 million trainable parameters. We have obtained very similar qualitative results using a larger transformer structure with 48 layers, embedding dimension 1600, and 25 attention heads, which gives approximately 1.5 billion trainable parameters.

Each transformer model is trained with the standard next-token prediction objective using a cross-entropy loss function. 
Given a token sequence $a_1,\dots,a_N$,
the model estimates $p_{\theta}(a_t|a_{<t})$, where $a_{<t}$ denotes the sequence of tokens preceding $a_t$, and $\theta$ denotes the model's parameters. Interpret $p_{\theta}(a|a_{<t})$ as the probability the model assigns to token $a$ to follow $a_{<t}$. The parameters $\theta$ are chosen through approximate minimization of the loss function
$$
L(\theta) = -\sum_{i=1}^N \log p_{\theta}(a_t|a_{<t}).
$$
For batched training over many sequences, this loss is summed over all non-padding tokens in the batch. 

For the experiments in Sections~\ref{s:diagnostics} to \ref{s:forecasts}, we use the base LOB setting \((P,V_1,V_2)=(3,5,2)\), and we train three models:

\begin{itemize}
    \item \textbf{SP}: trained on 500K shortest-path segments with geometric mean length 5 and cap 20.
    \item \textbf{RW-S}: trained on 10M random-walk segments with geometric mean length 20 and cap 60.
    \item \textbf{RW-L}: trained on 10M random-walk segments with geometric mean length 50 and cap 150.
\end{itemize}

We use 99\% of each dataset for training and reserve 1\% for heldout evaluation. All training is performed using mixed precision on a Supermicro AS-4124GO-NART+ server with eight NVIDIA A100-SXM4-80GB GPUs. The SP model is trained for 10 epochs. We run a validation every 5000 parameter updates and select the checkpoint with lowest validation loss. The RW models are trained for one epoch, and the final checkpoint is used. The training time is 4-6 hours for the SP model, and 1-2 days for the RW models.

In Section~\ref{s:scale_up}, we repeat the same pipeline on two larger LOB settings. One increases the depth capacity, using \((P,V_1,V_2)=(3,12,3)\), and the other increases the number of price levels, using \((P,V_1,V_2)=(8,2,2)\). For these scaling experiments, the training datasets are enlarged to two and three times the size of the corresponding base datasets, respectively, and the training time increases approximately proportionally with the amount of training data.

\section{Empirical-Kernel Baseline}
\label{s:empirical_baseline}

To separate errors caused by the training data from errors caused by the LLM's learned representation, we also construct a simple empirical-kernel baseline. This baseline has no hidden state or attention mechanism, it predicts the next event only based on the current LOB state by using empirical next-token frequencies in the training data. For each state \(s\) and event token \(a\) (including the \texttt{end} token), let \(N_{\mathrm{train}}(s,a)\) be the number of times event \(a\) appears immediately after state \(s\) in the training data, and let \(N_{\mathrm{train}}(s)=\sum_{a'\in\mathcal A} N_{\mathrm{train}}(s,a')\) be the total number of observed outgoing events from \(s\),
using the RW-L dataset. The empirical-kernel baseline predicts
\[
\widehat p_{\mathrm{emp}}(a\mid s)
=
\frac{N_{\mathrm{train}}(s,a)}
{N_{\mathrm{train}}(s)}
\]
For each LOB setting, the baseline is trained on the same dataset as the RW-L model. In our experiments, more than 99.8\% of evaluation states are observed in the training data and baseline results are computed over these covered states. This simple model provides a useful comparison: if this empirical-kernel baseline does not exhibit the same errors we find in the LLM, then those patterns are unlikely to be a consequence of the training data. Instead, they arise from the representation the LLM learns.

\section{Model Diagnostics}
\label{s:diagnostics}

\subsection{Valid Traversal Test}
\label{s:valid}

As a first test of the trained LLM's understanding of the LOB, we evaluate its ability to generate (only) valid sequences of tokens. We prompt the model with a heldout pair 
$(s_0, s_g)$. Recall that state IDs are themselves tokens, so we can evaluate the model's next-token distribution $p_{\theta}(\cdot|s_0, s_g)$
and choose token $\hat{a}_1$ to maximize the next-token probability. 
We then choose $\hat{a}_2$ to maximize $p_{\theta}(\cdot|s_0,s_g,\hat{a}_1)$
and so on to generate a sequence
$s_0,s_g, \hat{a}_1, \hat{a}_2, \cdots, \hat{a}_T$, \texttt{end}. We refer to this as greedy generation.

The generated trajectory is valid if every generated event is valid under the true LOB
transition rules and the terminal token appears only when the current state equals the target. Formally,
(recalling $\mathcal{A}$ and $T(\cdot,\cdot)$ from the end of Section~\ref{s:events})
starting from $\hat{s}_0=s_0$, each generated event must satisfy
$$
\hat{a}_t\in\mathcal{A}(\hat{s}_{t-1}),
\quad \hat{s}_t = T(\hat{s}_{t-1},\hat{a}_t),
$$
and \texttt{end} is emitted only when $\hat{s}_T=s_g$. We evaluate the fraction of valid sequences for each model.

In the experiments below, this evaluation is carried out on the testing split of each dataset. In addition to the valid traversal rate, we also report two shortest-path diagnostics. The SP Rate is the fraction of generated trajectories that reach \(s_g\) using a shortest path in the LOB state graph. The Extra Steps statistic is the generated path length minus the shortest-path distance between \(s_0\) and \(s_g\), averaged over valid generated trajectories.

Success in generating (only) valid sequences suggests an understanding of LOB dynamics. In contrast, executing a market buy order at a higher price than the lowest resting sell order would suggest a lack of understanding; so would generating a cancellation of a nonexistent limit order. Using sequences of the form (\ref{seq}) requires substantially more than this basic understanding of LOB rules by requiring the LLM to steer the state to a fixed goal in generation.

Table~\ref{tab:merged_results} shows that that all three LLMs score very high on the valid traversal test, nearly always generating valid sequences. As expected, the baseline performs poorly on this task because it has no mechanism for goal-directed sequence generation. The \textit{SP Rate} column shows that 99.4\% of sequences generated by the SP model also achieve the shortest length. Interestingly, the RW models also tend to generate short sequences. In the RW-S model, 90.6\% of generated sequences achieve the shortest path to the goal, even though only 10.4\% of sequences in its training set have this property. Also, the generated sequences of RW-S model exceed the shortest path by only 0.1 steps on average, whereas in its training data the value is 16.3. The RW-L results in the table show a similar pattern. The fact that RW models generate shorter sequences to reach $s_g$ further suggests an understanding of the LOB dynamics, since their training sequences are not constructed to be the shortest paths.

\begin{table}[H]
\centering
\fontsize{7.5}{9}\selectfont
\renewcommand{\arraystretch}{0.95}
\caption{Diagnostic results}
\label{tab:merged_results}
\begin{tabular}{@{}lcccc@{}}
\toprule
\textbf{Model} & \textbf{Valid} & \textbf{SP} 
& \textbf{Extra} 
& \textbf{Compress.} \\
\textbf{} & \textbf{Traversal} & \textbf{Rate} 
& \textbf{Steps} 
& \textbf{Score} \\
\midrule
SP   & 99.4\% & 99.4\% (100\%)   & 0.0 (0.0)  & 61.1\% \\
RW-S & 98.5\% & 90.6\% (10.4\%)  & 0.1 (16.3) & 42.0\% \\
RW-L & 95.5\% & 83.7\% (4.2\%)   & 0.3 (43.8) & 48.0\% \\
Baseline   & 0.7\% & 0.7\% (4.2\%)   & 0.0 (43.8)  & 99.9\% \\
\bottomrule
\end{tabular}
\end{table}

\subsection{Compression Test}
\label{s:compression}

The compression test, introduced in \cite{vafa2024evaluating}, examines an LLM's understanding of the state of the real system. Given the current state, an LLM with a correct world model should not be influenced by the history leading to that state in generating future sequences.
The test in \cite{vafa2024evaluating} was formulated in a deterministic setting; we adapt it to our setting for consistency with our data generating process.

We draw an initial state $s_0$ from the stationary distribution $\pi$. We simulate forward a geometric number of steps using the true LOB dynamics and record the state reached $s_g$; this will serve as the goal state. We then generate event histories by simulating backwards from state $s_0$. By Bayes' rule, state $s_0$ is preceded by state $s'$ and event $a$, where $T(s',a)=s_0$, with probability proportional to $\pi(s')\mu(a|s')$. Repeating this backward-sampling process produces an event history; by running the procedure twice we produce two histories (valid token sequences) $h_1$ and $h_2$ that terminate in $s_0$. 

From the history $h_1$ we use the LLM to generate suffix sequences $z=(z_1,\dots,z_m)$ with $z_1$ drawn from $p_{\theta}(\cdot|h_1,s_g)$, $z_2$ drawn from $p_{\theta}(\cdot|h_1,s_g,z_1)$, and so on. If the LLM's world model aligns with the LOB state, any suffix $z$ generated from $h_1$ should be feasible (according to the LLM) following $h_2$. Given a threshold $\epsilon>0$, we say that a suffix $z$ is accepted under $h_2$ if $p_{\theta}(z_j|h_2,s_g,z_{<j})>\epsilon$. The compression score is the fraction of states $s$ in which all suffixes generated from one incoming history with probability $>\epsilon$ are also accepted under a second incoming history. A high compression score suggests that the LLM should understand, given the fixed current state, the set of feasible suffixes does not depend on the path by which the current state was reached.

In our experiments, we use 1000 sampled state-goal-history pairs and 30 suffixes per pair. The prefix length is sampled from the capped geometric segment distribution. We set $\epsilon=0.01$, but the results are not sensitive to this threshold choice.

The compression scores of all the three LLMs in Table~\ref{tab:merged_results} are low, whereas the baseline score is close to 100\%. For the RW models, in more than half the states the set of valid suffixes is deemed history-dependent, even though in the true LOB dynamics the set of valid suffixes is identical for two histories ending in the same state $s_0$. 

We are thus left with a mixed picture. The LLMs perform remarkably well in generating valid sequences, but they have failed to understand a key feature of the LOB dynamics.%
\footnote{We have also applied the probe test of \cite{hewitt2019designing} and the detour and distinction tests of \cite{vafa2024evaluating}. The RW models perform well on the probe test, which checks if a model's hidden state encodes information about the true state. They also perform well on detour and distinction tests, while the SP model does not. These tests are less directly relevant to our focus on forecast distributions so we do not discuss them further.}
We turn next to the implications of this misunderstanding.

\section{Evaluating LLM Forecast Distributions}
\label{s:forecasts}

We will show that deficiencies in LLMs' implicit world view of LOB dynamics translate into biased forecasts and spurious predictability. This analysis requires refining the valid traversal and compression tests to account for the stochastic nature of the LOB dynamics.

Under greedy generation, an LLM will achieve a perfect score on the valid traversal test so long as the event that maximizes $p_{\theta}(\cdot|a_{<t})$ is always a valid event in the state determined by $a_{<t}$; beyond that requirement, the probabilities $p_{\theta}(a|a_{<t})$ do not affect the test. Similarly, the compression test checks that the suffix distributions that follow histories $h_1$ and $h_2$ have common support, but the threshold is a hard cap, and it ignores the probabilities assigned to valid suffixes. In the setting of a deterministic model, these probabilities are not necessarily relevant; in our stochastic setting, they form an important part of the LOB's dynamics. We therefore develop stochastic counterparts to the validity and compression tests.

\subsection{Kernel-Level TV Distance}
\label{s:tv1}

Under our LOB dynamics, an event sequence $a_{1:H}$ of length $H$ following state $s$ has probability
\[
P_H^*(a_{1:H}\mid s)
=
\prod_{t=1}^{H}
\mu(a_t\mid s_{t-1}),
\]
where $s_0=s$ and $s_t=T(s_{t-1},a_t)$.
We want to compare this probability with the LLM's implied probability. As before, $p_{\theta}(a|h,s_g)$ denotes the LLM's probability of event $a$ given history $h$ and goal state $s_g$. Then the LLM's implied probability of the token sequence $a_{1:H}$ given $(h,s_g)$ is
$$
Q_{\theta,H}(a_{1:H}|h,s_g) = 
p_{\theta}(a_1|h,s_g)\cdots p_{\theta}(a_H|h,s_g,a_{1:H-1})
$$
To remove the condition on the goal state while maintaining consistency with the training-data generation process, we simulate $M_g$ independent LOB paths from state $s$, producing $M_g$ goal states $s_{g_j}$. The resulting LLM probability, conditional on history $h$, is then estimated by the average over goal states
\begin{equation}
Q_{\theta,H}(\cdot\mid h)
=
\frac{1}{M_g}
\sum_{j=1}^{M_g}
Q_{\theta,H}(\cdot\mid h,s_{g,j}).
\label{qmg}
\end{equation}
The conditional kernel-level total variation compares the model's goal-averaged \(H\)-step path distribution with the true \(H\)-step kernel:
\begin{equation}
\mathrm{TV}^{\mathrm{kernel}}_H(h,s)
=
\frac{1}{2}
\sum_{a_{1:H}}
\left|
Q_{\theta,H}(a_{1:H}\mid h)
-
P_H^*(a_{1:H}\mid s)
\right|.
\label{tv1}
\end{equation}
Here, we retain \(p_{\theta}(a|h,s_g)\) assigned to invalid tokens \(a \notin \mathcal{A}(s)\), which have zero probability under \(P_H^*(\cdot \mid s)\) and therefore contribute directly to the TV distance. We average (\ref{tv1}) over draws of $s$ from the LOB's stationary distribution $\pi$ and from histories $h$ of geometric length ending in state $s$, using the backward-sampling procedure described in Section~\ref{s:compression}.

The resulting quantity measures how well the LLM recovers the true event-token transition probabilities at horizon \(H\). The value lies in \([0,1]\), with smaller values indicating stronger agreement. Whereas the compression test basically checks if the two distributions have the same support, $\mathrm{TV}^{\mathrm{kernel}}_H$ checks if they are close.

If the two distributions are not close, then there will be states in which the LLM predicts some events (e.g., a buy order at a particular price) has high probabilities while in fact they should not. We will illustrate this point with an example.

\subsection{Example 1: Kernel-Level TV}

With the notation of Figure~\ref{f:lob}, suppose the current state is \(s=(\ell_a,h_b,v_1,v_2,v_3)=(3,4,2,2,2)\). There are two limit sell orders at each price $p_1$, $p_2$, and $p_3$, and there are no limit buy orders.
Suppose this state was reached from state (2,4,4,1,0)
through the sequence of events and states in Table~\ref{tab:kernel_tv_prefix_example}.

\begin{table}[H]
\centering
\fontsize{7.5}{9}\selectfont
\caption{Prefix used in kernel-level TV}
\label{tab:kernel_tv_prefix_example}
\begin{tabular}{@{}clc@{}}
\toprule
\textbf{Step} & \textbf{Event} & \textbf{Resulting State} \\
\midrule
0 & Start & \((2,4,4,1,0)\) \\
1 & \texttt{ask\_p3\_2} & \((3,4,4,1,2)\) \\
2 & \texttt{bid\_p2\_1} & \((3,4,4,1,1)\) \\
3 & \texttt{ask\_p2\_1} & \((3,4,4,2,1)\) \\
4 & \texttt{w\_ask\_p1\_2} & \((3,4,2,2,1)\) \\
5 & \texttt{ask\_p3\_1} & \((3,4,2,2,2)\) \\
\bottomrule
\end{tabular}
\end{table}

In the current state \(s\) there are 20 valid events. The true one-step event kernel assigns probability \(1/20=0.05\) to each of them. The model's implied distribution is calculated using (\ref{qmg}), and the resulting one-step kernel-level TV is 0.665 in this example. 

To illustrate why the value is so large, Table~\ref{tab:kernel_tv_token_example} shows the five most likely events according to the model. The LLM assigns nearly a 50\% chance $(0.245+0.241)$ to the next event being a cancellation at price $p_1$, although the actual probability is only 10\%. It also overestimates the probability of a limit sell order at $p_1$ or $p_2$. 

In this example, the LLM's forecasts are severely biased. The errors illustrated in Table~\ref{tab:kernel_tv_token_example} would not prevent the LLM from attaining a perfect score on the valid traversal test, since that test is simply too coarse to quantify errors in forecasting distributions, which is why we introduced the kernel-level TV measure.

\begin{table}[H]
\centering
\fontsize{7.5}{9}\selectfont
\caption{Next-event probabilities in kernel-level TV}
\label{tab:kernel_tv_token_example}
\begin{tabular}{@{}lccc@{}}
\toprule
\textbf{Action Token} &
\textbf{Next State} &
\textbf{True Prob.} &
\textbf{LLM's Prob.} \\
\midrule
\texttt{w\_ask\_p1\_2} & \((3,4,0,2,2)\) & 0.050 & 0.245 \\
\texttt{w\_ask\_p1\_1} & \((3,4,1,2,2)\) & 0.050 & 0.241 \\
\texttt{ask\_p2\_2} & \((3,4,2,4,2)\) & 0.050 & 0.190 \\
\texttt{ask\_p1\_1} & \((3,4,3,2,2)\) & 0.050 & 0.132 \\
\texttt{ask\_p1\_2} & \((3,4,4,2,2)\) & 0.050 & 0.106 \\
\bottomrule
\end{tabular}
\end{table}

\subsection{History-Level TV Distance}
\label{s:history}

\begin{figure}[H]
    \centering
    \includegraphics[width=0.45\textwidth]{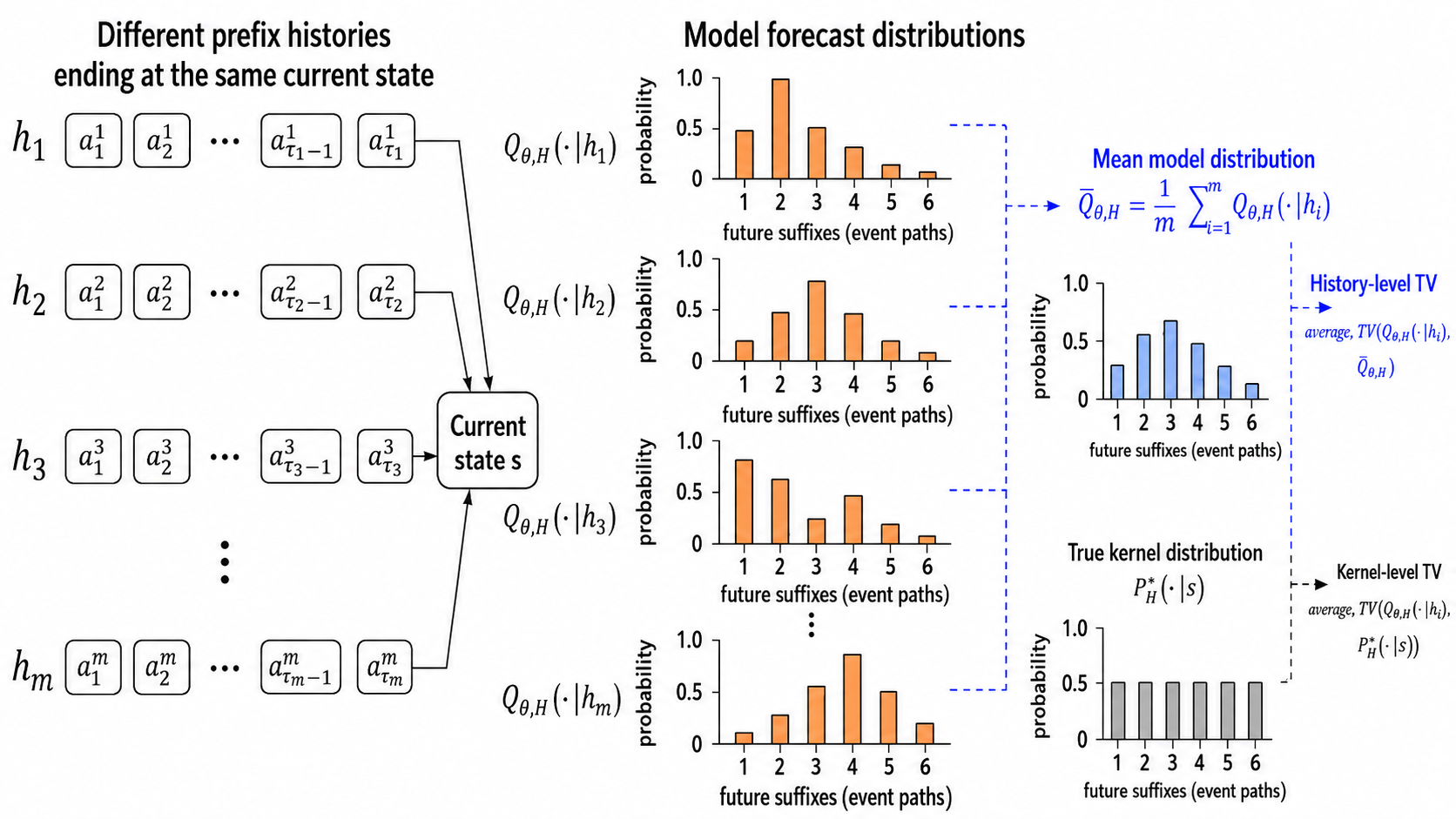}
    \caption{Illustration of history- and kernel-level TV}
    \label{f:TV}
\end{figure}

The kernel-level TV is essentially a measure of bias: it measures systematic errors in the LLM's forecast distribution when compared with the true dynamics. On the other hand, we also introduce the history-level TV, which measures whether the LLM gives different predictions for different histories ending at the same current state (see Figure \ref{f:TV}). For each current state \(s\), suppose we observe histories
$h_1,\ldots,h_m$, all ending at $s$.
Define the within-state mean model distribution \(\bar Q_{\theta,H}^{s}
=
\frac{1}{m}
\sum_{i=1}^{m}
Q_{\theta,H}(\cdot\mid h_i)\), then the history-level TV for one history $h_i$ and current state $s$ is
\begin{equation}
\mathrm{TV}^{\mathrm{history}}_H(h_i,s)
=
\frac{1}{2}
\sum_{a_{1:H}}
\left|
Q_{\theta,H}(a_{1:H}\mid h_i)
-
\bar Q_{\theta,H}^{s}(a_{1:H})
\right|.
\label{tv2}
\end{equation}
As with $\mathrm{TV}^{\mathrm{kernel}}_H$, we average this measure over draws of $s$ from $\pi$ and histories $h_i$ ending in $s$. This measure quantifies dispersion: the larger the value, the more the LLM's forecast is erroneously influenced by past history, which violates the Markov property.

In the experiments, the kernel-level TV is computed from 1000 sampled state-prefix pairs. For the history-level TV, it is computed from 200 sampled current state, and \(m=10\) prefixes for each state. The LLM's implied probability \(Q_{\theta,H}(a_{1:H}|h,s_g)\) for each state-prefix pair is averaged over \(M_g = 32\) sampled future goals.

Table~\ref{tab:rollout_tv_results} shows the results for our two TV measures across all models and three forecast horizons. The results are more relevant for the RW models. At $H=1$, the TV measures are rather small, but they clearly increase at longer forecasting horizons as error accumulates, which is a pattern we expect to see continue for larger $H$. Moreover, the variances of both TV measures are very small, indicating that the LLMs consistently exhibit forecast errors and excessive dependence on past events across all states.

\begin{table}[H]
\centering
\fontsize{7.5}{9}\selectfont
\caption{Multi-step forecast TV distances}
\label{tab:rollout_tv_results}
\begin{tabular}{@{}llccc@{}}
\toprule
\textbf{Model} &
\textbf{TV Type} &
$H=1$ & $H=2$ & $H=3$ \\
\midrule
\multirow{2}{*}{SP}
& Kernel-level  & 0.46 & 0.72 & 0.84 \\
& History-level & 0.33 & 0.51 & 0.59 \\
\midrule
\multirow{2}{*}{RW-S}
& Kernel-level  & 0.11 & 0.19 & 0.28 \\
& History-level & 0.07 & 0.14 & 0.23 \\
\midrule
\multirow{2}{*}{RW-L}
& Kernel-level  & 0.08 & 0.15 & 0.21 \\
& History-level & 0.06 & 0.11 & 0.16 \\
\midrule
\multirow{2}{*}{Baseline}
& Kernel-level  & 0.00 & 0.01 & 0.01 \\
& History-level & 0.00 & 0.00 & 0.01 \\
\bottomrule
\end{tabular}
\end{table}

\subsection{Example 2: History-Level TV}
\label{s:ex_tv2}

To illustrate the history-level TV measure, we consider two different histories that reach the same current state. The shared current state is
$s=(\ell_a,h_b,v_1,v_2,v_3)=(1,3,5,0,2)$. The two event sequences leading to this state are shown in Appendix~\ref{s:two_prefix}, Table~\ref{tab:history_tv_prefix_summary}.

We calculate $Q_{\theta,H}(\cdot|h_i)$, $i=1,2$, as in (\ref{qmg}) for the two histories, using the same sampled goal states $s_{g,j}$; this ensures the differences in $Q_{\theta,H}(\cdot|h_i)$ are due only to differences in the histories $h_i$, which should be irrelevant given the Markov property. The history-level TV distance (\ref{tv2}) using these histories is 0.385 in this example.

Table~\ref{tab:history_tv_token_example} provides more detail by showing the model-based probabilities of certain events following the two histories. If the LLM's implicit world view of the LOB recognized the current state, the last two columns would be identical. Instead, we see that the LLM assigns a rather large probability (0.278) to a limit buy order at price $p_2$ following $h_1$ and a rather large probability (0.313) of a limit sell order at price $p_3$ following $h_2$.

These are examples of spurious predictability: the LLM operates as if the history leading to the current state helps predict the next event, even though there should be no such predictability in our Markov-generated training data.

\begin{table}[H]
\centering
\fontsize{7.5}{9}\selectfont
\caption{Next-event probabilities in history-level TV}
\label{tab:history_tv_token_example}
\begin{tabular}{@{}lccc@{}}
\toprule
\textbf{Event Token} &
\textbf{Next State} &
\textbf{\(Q_{\theta,1}(a\mid h_1)\)} &
\textbf{\(Q_{\theta,1}(a\mid h_2)\)} \\
\midrule
\texttt{bid\_p2\_2} & \((1,2,5,2,2)\) & 0.278 & 0.002 \\
\texttt{ask\_p3\_2} & \((1,4,5,0,0)\) & 0.056 & 0.313 \\
\texttt{m\_ask\_2} & \((1,4,5,0,0)\) & 0.047 & 0.233 \\
\texttt{w\_bid\_p3\_2} & \((1,4,5,0,0)\) & 0.028 & 0.199 \\
\texttt{ask\_p2\_2} & \((2,3,5,2,2)\) & 0.053 & 0.208 \\
\texttt{ask\_p2\_1} & \((2,3,5,1,2)\) & 0.128 & 0.002 \\
\bottomrule
\end{tabular}
\end{table}

\subsection{Regression Test}
\label{s:regression}

A low compression score and a high history-level TV score indicate that the LLM has failed to understand the state of the LOB. A direct way to see the implications of this failure is to run a forecasting regression on features of the event history, which should again have no predictive power under the Markov property.

For our forecasting analysis, we group our target future events into four categories:
\begin{itemize}
    \item Buy orders: limit buy and market buy orders.
    \item Cancel orders: all cancellation orders.
    \item Market orders: all market orders.
    \item Big orders: orders with size greater than \(V_2/2\).
\end{itemize}
These categories correspond to economically meaningful order-flow quantities. Buy orders capture directional buy pressure. Cancel orders capture liquidity withdrawal from the book. Market orders capture aggressive liquidity-taking behavior. Big orders capture large-flow intensity. By studying the model’s forecasts for these quantities, we can identify which forms of financial predictability the LLM has incorrectly learned.

For each sampled visit, we compute the model's implied category ratios over \(H\)-step event paths. For a category \(c\), define
\[
r_c(a_{1:H})
=
\frac{1}{H}
\sum_{t=1}^{H}
\mathbf 1\{a_t\in c\}.
\]
The true expected category ratio is \(\mathbb E_{P_H^*}[r_c]\), and the model-implied expected category ratio is
\(\mathbb E_{Q_{\theta,H}}[r_c]\), with
\(P_H^*\) and \(Q_{\theta,H}\) as defined in Section~\ref{s:tv1}. We use the difference between these two quantities as the regression target:
\[
B_{c,H}(h,s)
=
\mathbb E_{Q_{\theta,H}}[r_c]
-
\mathbb E_{P_H^*}[r_c].
\]
It is important to note that the raw model's implied ratio
\(\mathbb E_{Q_{\theta,H}}[r_c]\) would not be the right object for detecting spurious predictability or miscalibration, because the true ratio
\(\mathbb E_{P_H^*}[r_c]\) should itself depend on the current state \(s\). Since different LOB states have different feasible actions and liquidity conditions, the true probabilities of buy orders, cancellations, market orders, and big orders may not be constant across states. For example, a state with little ask-side liquidity or a wide spread can have a different true distribution over future order types than a balanced, liquid state.

The ratio bias \(B_{c,H}(h,s)\) removes this true state-dependent component. It measures whether the LLM overpredicts or underpredicts a future order-flow quantity relative to the correct Markov benchmark of the true LOB dynamics.

We clarify the interpretation of the regression coefficients. Under the Markov LOB dynamics, the history \(h\) should have no incremental predictive content once the current state \(s\) is given. Significant prefix-history coefficients therefore reveal spurious predictability: the model assigns forecasting value to historical order flow that should be irrelevant. On the other hand, if current-state features predict \(B_{c,H}(h,s)\), this does not violate the Markov property directly, since both \(\mathbb E_{Q_{\theta,H}}[r_c]\) and \(\mathbb E_{P_H^*}[r_c]\) are allowed to depend on the current LOB state. However, because the target has already subtracted the true Markov benchmark, such coefficients indicate a state-dependent model miscalibration: the LLM is systematically too optimistic or pessimistic about future order flow in certain types of book configurations.

Based on that, for each category \(c\) and horizon \(H\in\{1,2,3\}\), we run an OLS regression,
\[
B_{c,H}(h,s)
=
\alpha
+
\beta_s^\top F_s(s)
+
\beta_h^\top F_h(h)
+
\varepsilon,
\]
where \(F_s(s)\) contains current-state features and \(F_h(h)\) contains prefix-history features. This gives 12 regression equations per model.

Here, we use \(D\) to represent the depth of the LOB. Recall the state definition \(s=(\ell^a,h^b,v_1,\dots,v_P)\) in Section~\ref{s:lob}, we have:
\[
D^{\mathrm{ask}}(s)
=
\sum_{i=1}^{\ell^a(s)} v_i(s), \quad
D^{\mathrm{best ask}}(s)=v_{\ell^a}(s).
\]
With similarly for bid depth, we also define
\(
D^{\mathrm{total}}(s)
=
D^{\mathrm{ask}}(s)+D^{\mathrm{bid}}(s).
\)
The current-state feature vector \(F_s(s)\) includes:
\begin{itemize}
    \item Current depth imbalance:
    \[
    \frac{
    D^{\mathrm{bid}}(s)-D^{\mathrm{ask}}(s)
    }{
    D^{\mathrm{bid}}(s)+D^{\mathrm{ask}}(s)
    }.
    \]
    This measures whether displayed liquidity is concentrated on the bid side or the ask side, which is a standard proxy for pressure at the book level.

    \item Current spread width: 
    \(
    \ell^a(s) - h^b(s)
    \). A wider spread corresponds to lower immediacy and liquidity.

    \item Current total depth: \(D^{\mathrm{total}}(s)\). This is the total displayed liquidity on the bid and ask sides. A deeper book is typically more resilient to incoming orders.

    \item Current best-depth imbalance:
    \[
    \frac{
    D^{\mathrm{best bid}}(s)-D^{\mathrm{best ask}}(s)
    }{
    D^{\mathrm{best bid}}(s)+D^{\mathrm{best ask}}(s)
    }.
    \]
    This measures the imbalance at the top of the book, focusing on the most immediately executable liquidity.
\end{itemize}

The prefix-history feature vector \(F_h(h)\) includes:
\begin{itemize}
    \item Prefix length: This is the number of past events in the segment. Under the Markov setting, segment age should not matter once the current state is known.

    \item Prefix buy ratio: This is the fraction of past events that are buy orders, which measures buy-side order-flow intensity.

    \item Prefix take ratio: This is the fraction of past events that are market orders or crossing-limit orders, which (partially) execute immediately and consume liquidity from the opposite side of the LOB. It measures the order-flow aggressiveness.

    \item Prefix cancel ratio:  This is the fraction of past events that are cancellations. It measures liquidity withdrawal and reflects active quote management.

    \item Prefix rest-order depletion:
    \[
    \sum_{t=1}^{|h|}
    \mathbf 1
    \left\{
    D^{\mathrm{total}}(s_t)
    <
    D^{\mathrm{total}}(s_{t-1})
    \right\},
    \]
    where \(s_{t-1}\) is the state before event \(a_t\), and \(s_t\) is the state after event \(a_t\). This counts how often events in the prefix reduce the total depth of the LOB, which captures liquidity removal through active trades or cancellations.

    \item Prefix event entropy: This measures the diversity of event types in the prefix. Low entropy corresponds to a concentrated, repetitive order-flow history, while high entropy corresponds to more varied flow.
\end{itemize}

\begin{figure}[H]
    \centering
    \includegraphics[width=0.45\textwidth]{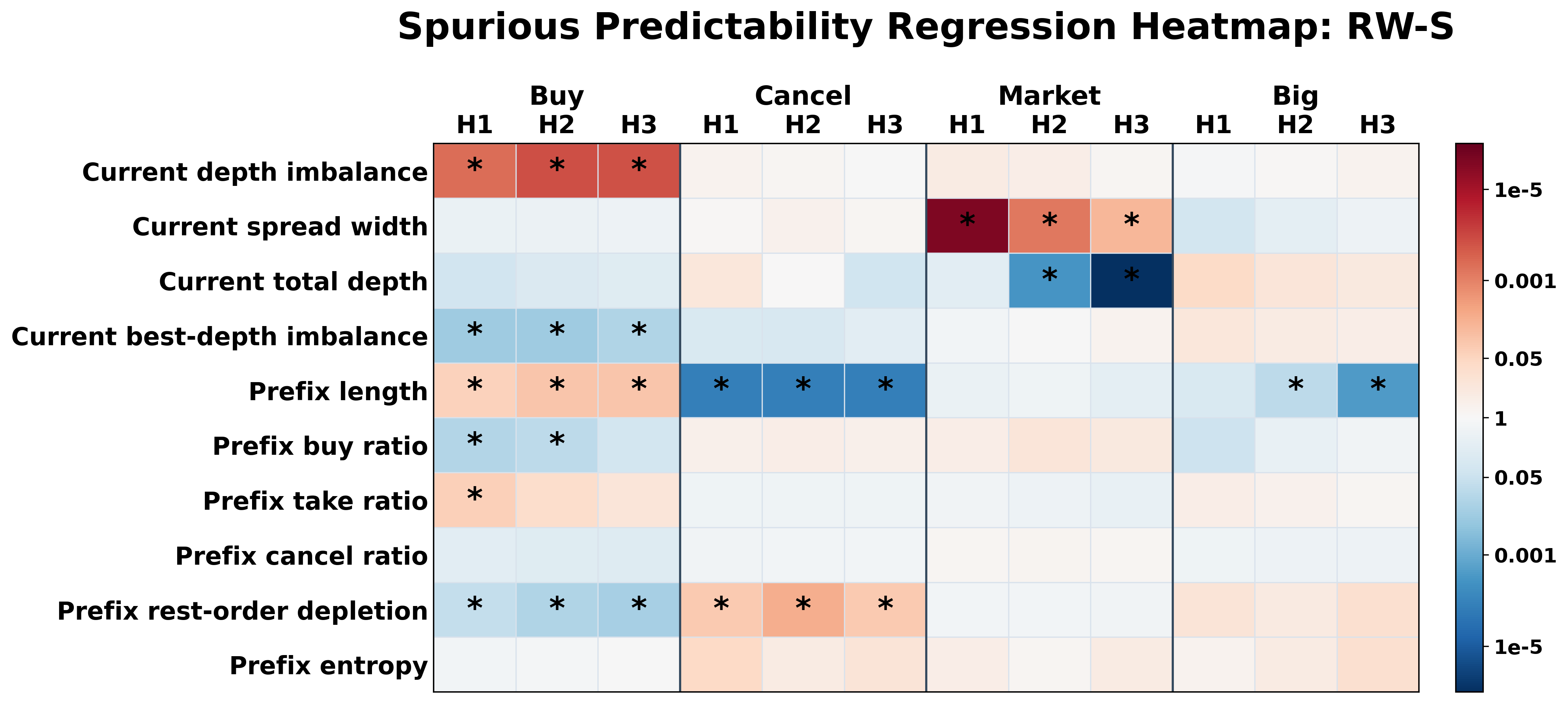}
    \caption{Regression results for RW-S}
    \label{f:regression-rws}
\end{figure}

Figure~\ref{f:regression-rws} shows results for the RW-S model. 
Each column corresponds to a separate regression for a given event category and forecasting horizon. 
The first four rows are state features, and the remaining six rows are prefix-history features. 
Red cells indicate positive coefficients, while blue cells indicate negative coefficients. 
A star marks coefficients that are statistically significant at the 5\% level%
\footnote{We use the 5\% significance level as a standard threshold and to provide a clear visual illustration. The results should be interpreted with the usual caution regarding multiple testing. Even under more conservative thresholds, like the Bonferroni correction, many features remain statistically significant.}
, and darker cells in either color indicate smaller \(p\)-values.

From the upper-left corner, we see that the LLM systematically overestimates the probability of a buy order in states with larger depth imbalance, and market orders in states with larger spreads,  while it underestimates market orders in states with greater depth for \(H=2,3\). These suggest that the model has state-dependent miscalibrations, that its forecast bias are wrongly related to current liquidity conditions. More importantly, in eight out of the twelve regressions, one or more prefix-history features appear to be statistically significant. For example, a higher historical buy ratio predicts a lower future buy probability, resembling a reversal signal in order flow; while greater rest-order depletion shifts the model's forecasts away from buy orders and toward cancellations, resembling a liquidity-withdrawal signal. Yet under the Markov property, these history features should be irrelevant to future events conditional on the current state. In other words, the predictability found by the LLM is spurious.

\begin{figure}[H]
    \centering
    \includegraphics[width=0.45\textwidth]{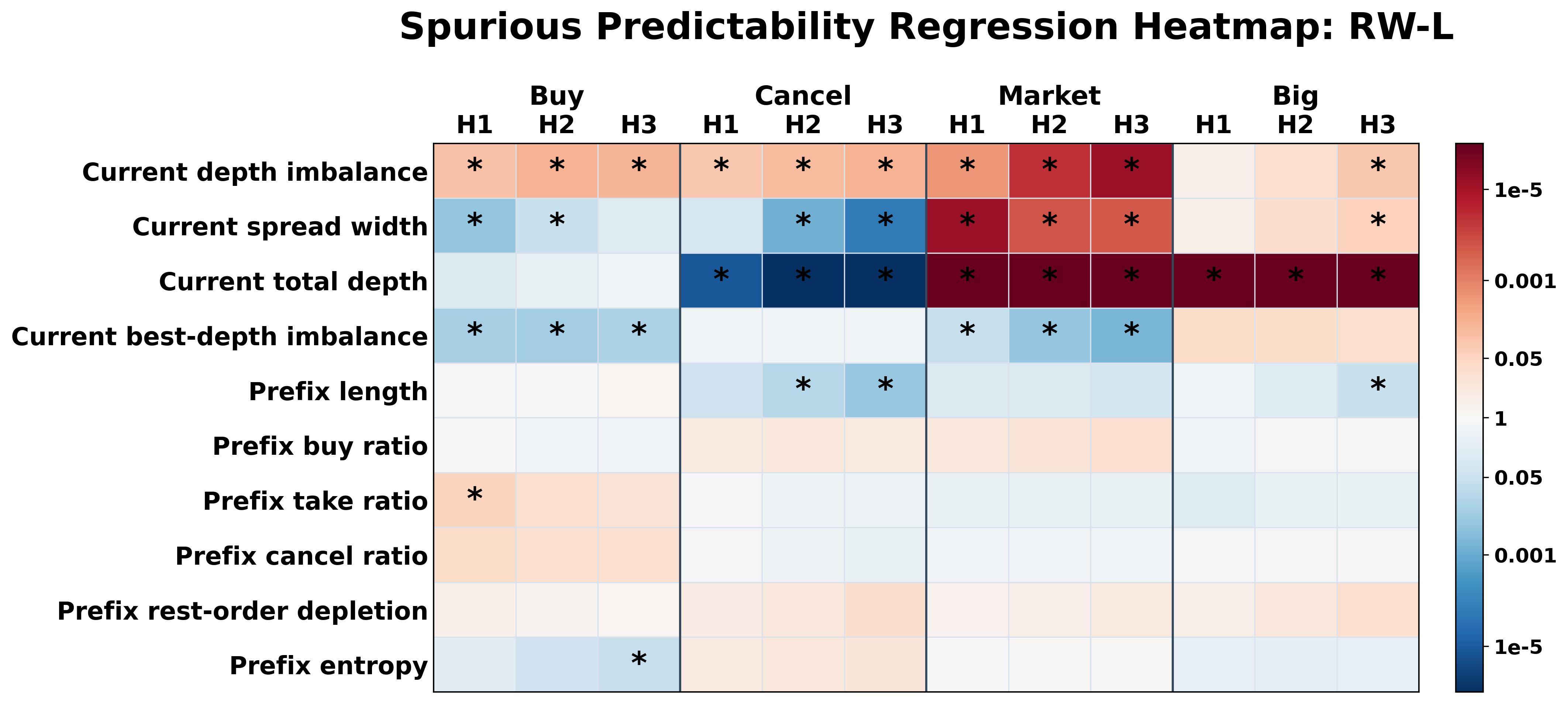}
    \caption{Regression results for RW-L}
    \label{f:regression-rwl}
\end{figure}

The results for RW-L in Figure~\ref{f:regression-rwl} are broadly similar.

\section{Scaling to Larger LOB Settings}
\label{s:scale_up}

As introduced in Section~\ref{s:training}, the base setting uses \((P,V_1,V_2)=(3,5,2)\), which gives a finite LOB space with 756 states. To check whether the main findings are specific to small environments, we repeat the same pipeline on two larger settings: \((P,V_1,V_2)=(3,12,3)\) and \((P,V_1,V_2)=(8,2,2)\). These settings substantially expand the number of reachable LOB states, to 8281 and 41553 states respectively.

Table~\ref{tab:merged_all_lob_results} shows that the qualitative conclusions are stable across these larger environments. The LLMs maintain high valid traversal rates and also tend to generate short sequence. However, the compression scores decline as the LOB environment becomes larger, especially for the RW models. The multi-step TV results point in the same direction, as kernel-level and history-level TV distances both increase in the larger settings. However, the baseline performs stably well. This suggests that the LLM's ability to represent the correct Markov state abstraction weakens as the LOB becomes more complex. Thus, the scaling experiments provide direct support for our claim that the issues we document would likely be more severe in larger settings, especially when LLMs are applied to real-market LOB data with larger price levels and size limits, and richer order-flow patterns.

\begin{table*}
\centering
\fontsize{7.5}{9}\selectfont
\caption{Diagnostic and multi-step forecast results for larger LOB settings}
\label{tab:merged_all_lob_results}
\begin{tabular}{@{}lc lcccc ccc ccc@{}}
\toprule
\multirow{2}{*}{\textbf{\makecell{LOB\\Setting}}} &
\multirow{2}{*}{\textbf{\makecell{\# of\\States}}} &
\multirow{2}{*}{\textbf{Model}} &
\multirow{2}{*}{\textbf{\makecell{Valid\\Traversal}}} &
\multirow{2}{*}{\textbf{\makecell{SP\\Rate}}} &
\multirow{2}{*}{\textbf{\makecell{Compress.\\Score}}} &
\multicolumn{3}{c}{\textbf{Kernel-level TV}} &
\multicolumn{3}{c}{\textbf{History-level TV}} \\
\cmidrule(lr){7-9}
\cmidrule(lr){10-12}
& & & & & & $H=1$ & $H=2$ & $H=3$ & $H=1$ & $H=2$ & $H=3$ \\
\midrule

\multirow{4}{*}{\makecell{$(P,V_1,V_2)$\\$=(3,12,3)$}}
& \multirow{4}{*}{8281}
& SP
& 99.4\% & 99.4\% (100\%) & 47.0\%
& 0.64 & 0.86 & 0.93 & 0.34 & 0.45 & 0.57 \\
&
& RW-S
& 97.8\% & 88.6\% (11.5\%) & 29.0\%
& 0.14 & 0.22 & 0.35 & 0.10 & 0.13 & 0.26 \\
&
& RW-L
& 94.1\% & 81.2\% (4.0\%) & 34.0\%
& 0.13 & 0.19 & 0.32 & 0.08 & 0.14 & 0.18 \\
&
& Baseline
& 0.3\% & 0.3\% (4.0\%) & 99.8\%
& 0.00 & 0.02 & 0.01 & 0.01 & 0.00 & 0.01 \\

\midrule

\multirow{4}{*}{\makecell{$(P,V_1,V_2)$\\$=(8,2,2)$}}
& \multirow{4}{*}{41553}
& SP
& 97.1\% & 96.8\% (100\%) & 53.6\%
& 0.61 & 0.79 & 0.92 & 0.39 & 0.56 & 0.62 \\
&
& RW-S
& 98.6\% & 90.2\% (10.1\%) & 23.0\%
& 0.17 & 0.24 & 0.38 & 0.12 & 0.16 & 0.27 \\
&
& RW-L
& 99.3\% & 83.3\% (6.0\%) & 24.0\%
& 0.16 & 0.20 & 0.31 & 0.11 & 0.18 & 0.23 \\
&
& Baseline
& 0.4\% & 0.4\% (6.0\%) & 99.9\%
& 0.00 & 0.01 & 0.02 & 0.00 & 0.01 & 0.01 \\

\bottomrule
\end{tabular}
\end{table*}

\section{Conclusion}
\label{s:conclusion}

Our investigation contributes to the study of world models implicit in LLMs, with a specific focus on understanding the operation of a financial market --- a limit order book.
We have shown that an LLM trained on LOB event sequences may perform well in generating valid sequences yet fail to understand the dynamics of the LOB. This deficiency leads to biased forecasts and spurious predictability in LOB events. We introduce new tests to measure these effects.

To train an LLM from scratch, we keep our LOB settings small. A larger LOB would likely require vastly more training data to perform well on our tests, consistent with our scaling-up experiments. By using synthetic data, we are able to evaluate LLM forecasts in a setting without predictability. Events in an actual LOB may exhibit some predictability, but an LLM's ability to provide reliable forecasts is suspect if it finds predictability where none exists.

\bibliography{LOB_LLM_refs}

@inproceedings{assefa2020generating,
  title={Generating synthetic data in finance: opportunities, challenges and pitfalls},
  author={Assefa, Samuel A and Dervovic, Danial and Mahfouz, Mahmoud and Tillman, Robert E and Reddy, Prashant and Veloso, Manuela},
  booktitle={Proceedings of the first ACM international conference on AI in finance},
  pages={1--8},
  year={2020}
}

@book{cartea2015algorithmic,
  title={Algorithmic and high-frequency trading},
  author={Cartea, {\'A}lvaro and Jaimungal, Sebastian and Penalva, Jos{\'e}},
  year={2015},
  publisher={Cambridge University Press}
}

@article{guan2023leveraging,
  title={Leveraging pre-trained large language models to construct and utilize world models for model-based task planning},
  author={Guan, Lin and Valmeekam, Karthik and Sreedharan, Sarath and Kambhampati, Subbarao},
  journal={Advances in Neural Information Processing Systems},
  volume={36},
  pages={79081--79094},
  year={2023}
}

@inproceedings{hazineh2023linear,
  title={Linear latent world models in simple transformers: A case study on Othello-GPT},
  author={Hazineh, Dean and Zhang, Zechen and Chiu, Jeffrey},
  booktitle={Socially Responsible Language Modelling Research},
  year={2023}
}

@inproceedings{hewitt2019designing,
  title={Designing and interpreting probes with control tasks},
  author={Hewitt, John and Liang, Percy},
  booktitle={Proceedings of the 2019 conference on empirical methods in natural language processing and the 9th international joint conference on natural language processing (emnlp-ijcnlp)},
  pages={2733--2743},
  year={2019}
}

@article{kong2024large,
  title={Large Language Models for Financial and Investment Management: Applications and Benchmarks},
  author={Kong, Yaxuan and Nie, Yuqi and Dong, Xiaowen and Mulvey, John M and Poor, H Vincent and Wen, Qingsong and Zohren, Stefan},
  journal={Journal of Portfolio Management},
  volume={51},
  number={2},
  pages={162--210},
  year={2024},
  publisher={Portfolio Management Research}
}

@article{li2023emergent,
  title={Emergent World Representations: Exploring a Sequence Model Trained on a Synthetic Task},
  author={Li, Kenneth and Hopkins, Aspen K and Bau, David and Vi{\'e}gas, Fernanda and Pfister, Hanspeter and Wattenberg, Martin},
  journal={ICLR},
  year={2023},
  publisher={ICLR}
}

@inproceedings{liutransformers,
  title={Transformers Learn Shortcuts to Automata},
  author={Liu, Bingbin and Ash, Jordan T and Goel, Surbhi and Krishnamurthy, Akshay and Zhang, Cyril},
  booktitle={The Eleventh International Conference on Learning Representations},
  year={2023}
}

@inproceedings{NagyFSLCZF23,
  author={Peer Nagy and Sascha Frey and Silvia Sapora and Kang Li and Anisoara Calinescu and Stefan Zohren and Jakob N. Foerster},
  title={Generative AI for End-to-End Limit Order Book Modelling: A Token-Level Autoregressive Generative Model of Message Flow Using a Deep State Space Network},
  year={2023},
  pages={91-99},
  url={https://doi.org/10.1145/3604237.3626898},
  booktitle={Proceedings of the 4th ACM International Conference on AI in Finance (ICAIF 2023)},
}

@inproceedings{toshniwal2022chess,
  title={Chess as a testbed for language model state tracking},
  author={Toshniwal, Shubham and Wiseman, Sam and Livescu, Karen and Gimpel, Kevin},
  booktitle={Proceedings of the AAAI Conference on Artificial Intelligence},
  volume={36},
  number={10},
  pages={11385--11393},
  year={2022}
}

@article{vafa2024evaluating,
  title={Evaluating the world model implicit in a generative model},
  author={Vafa, Keyon and Chen, Justin Y and Rambachan, Ashesh and Kleinberg, Jon and Mullainathan, Sendhil},
  journal={Advances in Neural Information Processing Systems},
  volume={37},
  pages={26941--26975},
  year={2024}
}

@inproceedings{vafa2025has,
  title={What Has a Foundation Model Found? Using Inductive Bias to Probe for World Models},
  author={Vafa, Keyon and Chang, Peter G and Rambachan, Ashesh and Mullainathan, Sendhil},
  booktitle={International Conference on Machine Learning},
  pages={60727--60747},
  year={2025},
  organization={PMLR}
}

\appendix

\section{Supporting Information}

\subsection{Shortest Paths Datasets Construction}
\label{s:shortestpath}

To generate a shortest-path training sequence (\ref{seq}), we first draw the goal state $s_g$ from the stationary distribution $\pi$, and we then run a "reverse breadth-first search (BFS)" on the LOB state transition graph. Firstly, we calculate the distance $d_{s_g}(s)$ from every state $s$ to $s_g$ using normal BFS.
Then, starting from $d_{s_g}({s_g})=0$, we work backwards from the endpoint using the fact that if there exists a valid event \(a\) such that \(T(s,a)=s'\), then $d_{s_g}(s')=d_{s_g}(s)-1$. We repeatedly choose an incoming event-labeled predecessor whose distance is exactly one larger than the current distance. After each reverse step, we stop with probability \(p_{\mathrm{stop}}\). Reversing the sampled event sequence gives a forward shortest path from \(s_{\mathrm{start}}\) to \({s_g}\). When multiple event-labeled edges satisfy the distance condition, uniform sampling is done at the event level.

\subsection{Two Prefixes in History-Level TV Example}
\label{s:two_prefix}

We show the two histories used in Section~\ref{s:ex_tv2} in Table~\ref{tab:history_tv_prefix_summary}.

\begin{table}[H]
\centering
\fontsize{7.5}{9}\selectfont
\caption{Two prefixes in history-level TV}
\label{tab:history_tv_prefix_summary}
\begin{tabular}{@{}clp{0.28\textwidth}@{}}
\toprule
\textbf{Prefix} & \textbf{Start State} & \textbf{Event Sequence} \\
\midrule
\(h_1\) &
\((2,3,2,2,1)\) &
\texttt{ask\_p1\_2},
\texttt{w\_bid\_p3\_1},
\texttt{ask\_p2\_2},
\texttt{ask\_p1\_1},
\texttt{bid\_p3\_2},
\texttt{bid\_p2\_1},
\texttt{m\_ask\_1},
\texttt{bid\_p3\_1},
\texttt{w\_bid\_p3\_2},
\texttt{ask\_p2\_1},
\texttt{ask\_p3\_2},
\texttt{bid\_p1\_1},
\texttt{bid\_p3\_2},
\texttt{m\_ask\_1},
\texttt{m\_bid\_2},
\texttt{m\_bid\_1},
\texttt{m\_bid\_2},
\texttt{bid\_p1\_1},
\texttt{ask\_p1\_2},
\texttt{bid\_p3\_2}
\\
\midrule
\(h_2\) &
\((1,2,3,1,2)\) &
\texttt{bid\_p3\_2},
\texttt{bid\_p2\_1},
\texttt{bid\_p2\_2},
\texttt{m\_bid\_2},
\texttt{ask\_p1\_1},
\texttt{w\_bid\_p3\_1},
\texttt{w\_bid\_p3\_1},
\texttt{ask\_p1\_2},
\texttt{w\_bid\_p3\_2},
\texttt{ask\_p3\_1},
\texttt{m\_ask\_2},
\texttt{bid\_p3\_1},
\texttt{ask\_p3\_1},
\texttt{w\_ask\_p1\_1},
\texttt{ask\_p2\_2},
\texttt{ask\_p1\_1},
\texttt{m\_ask\_1},
\texttt{ask\_p1\_1},
\texttt{m\_bid\_2},
\texttt{bid\_p3\_2}
\\
\bottomrule
\end{tabular}
\end{table}

\end{document}